\documentclass[twocolumn, 10pt, a4paper]{article}
\usepackage[margin=1in]{geometry}
\usepackage{authblk} % for author affiliations
\usepackage{amsmath, amssymb}
\usepackage{graphicx}
\usepackage{subcaption}
\usepackage[numbers]{natbib}
\usepackage{hyperref}
\usepackage{booktabs}
\usepackage{makecell}

\title{Engineering Regression Without Real-Data Training: \\ Domain Adaptation for Tabular Foundation Models Using Multi-Dataset Embeddings}

\author[1]{Lyle Regenwetter}
\author[1]{Rosen Yu}
\author[1]{Cyril Picard}
\author[1]{Faez Ahmed}
\affil[1]{Massachusetts Institute of Technology}

\date{} % No date, or use \date{\today}

\begin{document}
\maketitle

\begin{abstract}
Predictive modeling in engineering applications has long been dominated by bespoke models and small, siloed tabular datasets, limiting the applicability of large-scale learning approaches. Despite recent progress in tabular foundation models, the resulting synthetic training distributions used for pre-training may not reflect the statistical structure of engineering data, limiting transfer to engineering regression. We introduce TREDBench, a curated collection of 83 real-world tabular regression datasets with expert engineering/non-engineering labels, and use TabPFN 2.5’s dataset-level embedding to study domain structure in a common representation space. We find that engineering datasets are partially distinguishable from non-engineering datasets, while standard procedurally generated datasets are highly distinguishable from engineering datasets, revealing a substantial synthetic–real domain gap. To bridge this gap without training on real engineering samples, we propose an embedding-guided synthetic data curation method: we generate and identify ``engineering-like'' synthetic datasets, and perform continued pre-training of TabPFN 2.5 using only the selected synthetic tasks. Across 35 engineering regression datasets, this synthetic-only adaptation improves predictive accuracy and data efficiency, outperforming TabPFN 2.5 on 29/35 datasets and AutoGluon on 27/35, with mean multiplicative data-efficiency gains of 1.75x and 4.44x, respectively.
More broadly, our results indicate that principled synthetic data curation can convert procedural generators into domain-relevant ``data engines,'' enabling foundation models to improve in data-sparse scientific and industrial domains where real data collection is the primary bottleneck.

\end{abstract}

\section{Introduction}
Predictive modeling plays an increasingly important role in engineering design and decision-making. In many engineering workflows, evaluating candidate designs requires expensive simulations or physical experiments. Predictive surrogate models offer a practical alternative by approximating the relationship between design variables and performance metrics, enabling faster exploration of design spaces and more efficient optimization. However, building reliable predictive models in engineering remains challenging because datasets are often small, heterogeneous, and expensive to obtain.

As a result, data-driven predictive modeling in engineering has historically relied on bespoke models trained separately for each dataset or task. Classical machine learning methods such as Gaussian Processes, gradient boosting, and neural networks are commonly used, but each new dataset typically requires retraining, hyperparameter tuning, and careful model selection. This paradigm contrasts sharply with recent developments in other areas of applied AI. In fields such as natural language processing and computer vision, foundation models trained on large and diverse datasets—such as GPT-style large language models for text and vision foundation models like CLIP and diffusion models—have consolidated many previously distinct tasks into unified systems capable of generalizing across domains.

Many engineering predictive modeling tasks are naturally represented in tabular form, where rows correspond to candidate designs or experiments and columns represent design variables and performance metrics. A major barrier to foundation models in engineering is the scarcity and fragmentation of available datasets. Collecting large-scale training corpora comparable to those used for language or vision models is difficult because engineering data is often expensive to generate. For example, a dataset of a dozen vehicle crash tests can cost \$1 million to generate. 

Recent work in tabular machine learning has proposed an alternative approach: training models on large collections of procedurally generated synthetic datasets that approximate a broad distribution of prediction problems. Prior-data fitted networks (PFNs), such as TabPFN, learn predictive inference procedures by pretraining on millions of synthetic datasets and can subsequently perform predictions on new tasks using in-context learning.

These developments raise a fundamental question for engineering applications: \emph{Can a foundation model trained entirely on synthetic datasets achieve strong predictive performance on real engineering data?} Because machine learning models generally perform best on data drawn from distributions similar to their training data, this leads to a second question: \emph{how closely do procedurally generated datasets resemble real engineering datasets?} If synthetic data can approximate the statistical structure of engineering problems, then procedurally generated training pipelines may offer a scalable solution to the long-standing data scarcity in engineering machine learning.

In this work, we investigate this hypothesis through a systematic study of engineering tabular regression problems. We introduce \textbf{TREDBench}, a benchmark for Tabular Regression on Engineering Datasets. TREDBench is a curated collection of 35 engineering-related and 48 non-engineering tabular regression datasets. Using this benchmark, we analyze the distributional structure of engineering datasets and compare them to both non-engineering datasets and procedurally generated synthetic data used by tabular foundation models. 
% To enable such comparisons, we develop a set of dataset-level similarity metrics and dataset embeddings that allow entire datasets to be compared in a shared representation space.

\begin{figure*}[tb]
    \centering
    \includegraphics[width=\linewidth]{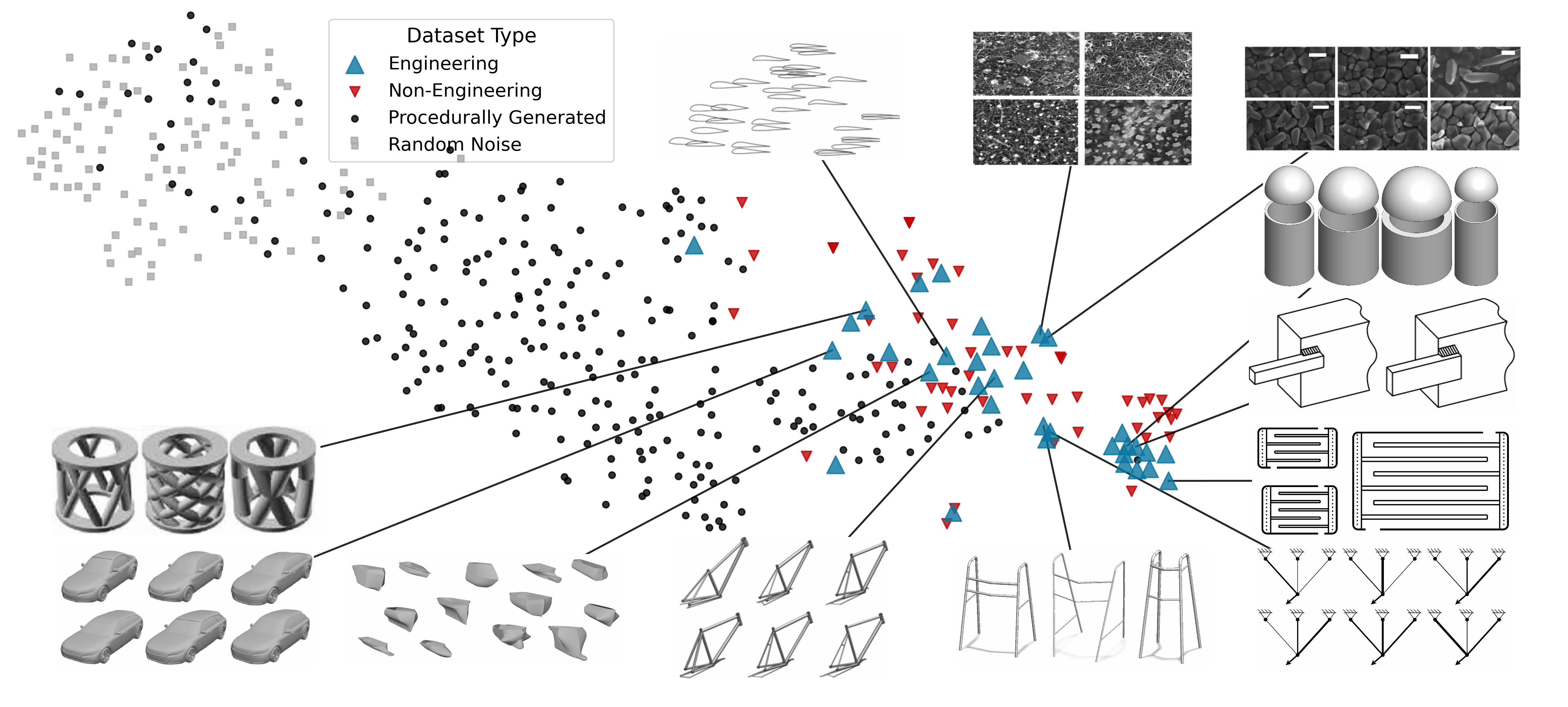}
    \caption{2D T-SNE embedding of TabPFN 2.5's 192-dimensional dataset embedding. Distributional differences between engineering and non-engineering data are apparent. A small proportion of the procedurally generated data overlaps with the engineering data. The procedurally generated data appears to splay off toward random noise, indicating that a significant amount of the procedurally generated training data may be too random to be realistic.}
    \label{fig:embedding}
\end{figure*}

Our analysis reveals that engineering datasets occupy a partially distinct region of dataset space, but that a subset of procedurally generated datasets closely resembles real engineering data, as illustrated in Figure~\ref{fig:embedding}. Leveraging this insight, we propose a method for selecting synthetic datasets that are most similar to engineering datasets and use them to fine-tune a tabular foundation model, TabPFN 2.5, without access to real engineering data. The resulting model achieves substantial improvements in predictive performance and data efficiency, outperforming strong tabular learning baselines—including both the original TabPFN 2.5 and the AutoGluon AutoML system—across the majority of the evaluated engineering problems.

Together, these results suggest that synthetic data may offer a viable path toward foundation models for engineering predictive modeling. By demonstrating that carefully selected procedurally generated datasets can substantially improve performance on real engineering problems, this work provides evidence that the longstanding data scarcity in engineering can be mitigated through principled synthetic data generation and selection. More broadly, our findings suggest that procedural generators can be transformed into domain-relevant ``data engines'' through principled synthetic data curation, enabling foundation models to improve in data-sparse scientific and industrial domains where real data collection is the primary bottleneck. Key contributions are summarized as follows:

%TRED-mark
%Engineering Tabular Regression Benchmark
\begin{itemize}
    \item \textbf{Engineering Tabular Benchmark}: We curate TREDBench, a collection of 83 tabular regression datasets, manually labeled as engineering vs. non-engineering by domain experts, and standardized for consistent small-data evaluation.
    \item \textbf{Quantifying the engineering–synthetic domain gap}: Using dataset-level embeddings and a dataset classifier in embedding space, we show (i) engineering vs. non-engineering real datasets are distinguishable (65.8\% accuracy) and (ii) procedurally generated synthetic vs. engineering datasets are highly distinguishable (96.8\%), indicating a substantial mismatch between standard procedural generators and real engineering data. 
    \item \textbf{Embedding-guided synthetic data curation for synthetic-only adaptation}: We propose an adaptation pipeline that generates 10,000 procedural synthetic regression datasets, embeds them, trains an embedding-space classifier, and selects the top 200 ``engineering-like'' synthetic datasets by predicted engineering probability—reducing synthetic-vs-engineering distinguishability to 88.9\%.
    \item \textbf{Synthetic-only adaptation of a tabular foundation model}: We propose an algorithm that performs continued pre-training of TabPFN 2.5 using only the selected synthetic datasets (no real engineering samples are used to update model weights), yielding consistent improvements on engineering regression tasks. By evaluating performance across training-set-size sweeps and reporting data-efficiency improvements, we show significant wins of our method over both the base TabPFN 2.5 and AutoGluon across 35 engineering datasets, along with 1.75x average data-efficiency.

\end{itemize}

\section{Background}
\subsection{Predictive Modeling in Engineering}

Engineering design increasingly relies on predictive models to guide decision-making. Evaluating candidate designs often requires expensive physics simulations or experiments, making it impractical to explore large design spaces directly. These predictive models (surrogates) provide an efficient alternative to resource-intensive simulations by approximating the mapping between design variables and system performance \cite{mu2025guide,ravi2025interpretable}. Common examples include Gaussian Processes, tree-based ensembles like XGBoost, and neural network architectures \cite{picard2024untrained,sajadi2025two,yousefpour2025simultaneous}. These models can accelerate design exploration, guide optimization, and identify feasible solutions before expensive evaluations are performed.

However, developing reliable predictive models remains challenging in engineering contexts. Datasets are often small, heterogeneous, and expensive to collect. Many engineering problems therefore operate in a low-data regime, where conventional machine learning models may struggle to generalize. Over the past decade, research has emphasized the need for curated datasets and improved data practices to enable data-driven engineering design \cite{ruiz2017data,rahman2019computer,ahmed2025design}. Efforts to generate synthetic datasets and large-scale design repositories have also been proposed to address these limitations \cite{lejeune2020mechanical,lee2023t,felten2025engibench,karri2025huver}. In parallel, physics-guided learning and model-free inference methods have been explored to integrate physical knowledge with data-driven approaches \cite{faroughi2024physics,haghighat2021physics,prume2023model}.

% Data scarcity, cyril's work \cite{picard2024untrained}, and anything else we deem relevant

\subsection{Tabular Foundation Models}
Engineering datasets are often represented in parametric, tabular form. Typically, each row corresponds to a candidate design or experiment, while each column represents a design parameter or measured performance metric. For many years, classical machine learning algorithms such as gradient boosting, random forests, and AutoGluon ensembles have been the dominant methods for tabular prediction tasks \cite{regenwetter2023framed,lee2025accessible}. In practice, these approaches often outperform deep neural networks on structured datasets \cite{shwartz2022tabular}. However, these models still require training a new model for each dataset, which can involve extensive hyperparameter tuning and model selection.

Recent work has begun exploring tabular foundation models (TFMs), which aim to learn predictive patterns across many datasets. Similar to large language models (LLMs), TFMs are trained on large collections of prediction tasks and can generalize to new datasets with minimal additional training \cite{pmlr-v235-van-breugel24a}. At inference time, similar to LLMs that learn to predict text conditioned on the context (the prompt), TFMs learn to predict posterior distributions conditioned on the in-context dataset. These developments suggest a new paradigm for predictive modeling in engineering: reusable models that can generalize across design problems without retraining.

%Benchmark efforts such as TabArena highlight the growing interest in evaluating such models across diverse tabular learning problems \cite{erickson2025tabarena}. These developments suggest a new paradigm for predictive modeling in engineering: reusable models that can generalize across design problems without retraining.

\subsection{Prior-Data Fitted Networks}
Prior-data fitted networks (PFNs) are state-of-the-art (SOTA) models that have advanced the field of TFMs. PFNs are pre-trained offline on millions of synthetic datasets generated from structural causal priors using a data-generating pipeline \cite{hollmann2023tabpfn,muller2022transformers,hollmann2025accurate}. Through this pretraining process, the transformer architecture learns to approximate Bayesian inference across a wide range of prediction tasks. At inference time, PFNs perform prediction using in-context learning. The model receives labeled examples and unlabeled samples simultaneously and directly outputs predictions in a single forward pass. This eliminates the need to train a new model for each dataset. 

TabPFN is a transformer-based PFN designed specifically for tabular data. While earlier iterations were limited to small datasets, the latest TabPFN v2 \cite{hollmann2025accurate} and v2.5 \cite{grinsztajn2025tabpfn} variants have scaled this power to datasets with up to 50,000 samples and 2,000 features. These models consistently outperform tuned gradient boosting methods and complex AutoML ensembles like AutoGluon in a fraction of the time on the leading tabular machine learning benchmark. 

% Relevant citations: \cite{vu2025adaptation,yu2025fast,yu2026fire}

These properties make PFNs particularly attractive for engineering design problems, where datasets are small and predictive models must be deployed quickly across different tasks. Recent work also suggests that TabPFN can be repurposed as a feature encoder for tabular data. In particular, the transformer representations produced during in-context inference form a highly separable embedding space for tabular instances, enabling downstream tasks such as visualization or lightweight predictive models built on top of these embeddings \cite{hollmann2025accurate,ye2025closer}.

\subsection{PFNs in Engineering}

Recent work has begun exploring PFNs for engineering problems. In particular, \citet{picard2024untrained} evaluated TabPFN across several canonical engineering design benchmarks and found that it often outperforms classical surrogate modeling approaches such as Gaussian processes and gradient boosting. However, performance varies substantially across problems: while TabPFN performs strongly on many engineering datasets, it performs poorly on others, including well-studied benchmarks such as the welded beam design problem. The causes of these inconsistencies remain unclear. One plausible explanation is that the procedurally generated training tasks used to pretrain PFNs may not fully capture the statistical structure of certain engineering datasets. Despite the growing interest in tabular foundation models for scientific and engineering applications, there has been little systematic investigation into how closely synthetic PFN training data resembles real engineering datasets, or how such generators might be adapted to better reflect domain-specific data distributions.

\section{Datasets}
This study aims to identify systematic differences between real engineering data, real non-engineering data, and procedurally generated synthetic data. We therefore source large collections of real data, then classify this data as engineering versus non-engineering.
We focus exclusively on tabular regression data. 

\subsection{Real Data}

\paragraph{Sourcing Tabular Regression Datasets.} To create a large pool of real-world tabular regression datasets, we selected three of the most popular tabular regression dataset collections from OpenML~\cite{vanschoren2014openml}: ``AutoML Benchmark'' (ID 269)~\cite{gijsbers2024amlb}, ``Tabular Benchmark'' (IDs 335, 336)~\cite{grinsztajn2022tree}, and ``OpenML CTR23'' (ID 353)~\cite{fischer2023openml}. We supplemented these datasets with imbalanced tabular regression datasets from~\cite{nejjar2024context}. Because these collections have relatively few engineering datasets, we specifically augmented our dataset pool with several small collections of engineering datasets from engineering design benchmarks~\cite{picard2024untrained, liang2021benchmarking, yu2025fast}. Finally, we included four datasets~\cite{narayanan2023data, narayanan2023data, elrefaie2024drivaernet, regenwetter2025bikebench} that were voluntarily contributed to the public data repository by the dataset creators and which we verified for rigor and integrity. A full list of datasets, their sources, and their labels is included in Appendix~\ref{Sec:dataset_details}.

\paragraph{Processing Real Datasets.} The full pool of datasets was then filtered according to a strict set of selection criteria. First, duplicate datasets were dropped. Next, categorical features were one-hot encoded. To cap computational cost and standardize the collection, a random set of 1024 datapoints from each dataset were selected. Datasets with fewer than 1024 samples underwent random duplication for embedding generation, but not for benchmarking, avoiding any data-leakage issue. In addition, datasets with more than 100 features were discarded. This decision was both to cap complexity and to accommodate PFN-style models with limited context constraints. For datasets with multiple regression labels, one label was selected at random. Each dataset's features and the selected label were independently scaled to mean 0 and variance 1.

\paragraph{Labeling Engineering Datasets.} Finally, datasets were manually classified as engineering-related or non-engineering-related by three human experts. Datasets classified into the engineering category primarily concern topics related to mechanical, material, chemical, and electrical engineering. Datasets classified into the non-engineering category concern a variety of topics including real-estate, economics, logistics, and culture.

\subsection{Synthetic Data}
Procedurally generated data is created using the procedure proposed in TabICL~\cite{qu2025tabicl}. To match the real datasets, each synthetic dataset is generated with 1024 datapoints and between 2 and 100 features. Like the real data, each synthetic dataset's features and label were independently scaled to mean 0 and variance 1. 

\subsection{Control Data} As a control, we created a collection of fully random datasets with between 2 and 100 features. Each feature and label was independently sampled from a standard Gaussian distribution. This dataset represents data with no `signal'--- where any relationship discerned between features and label are only attributable to spurious noise. 

\subsection{An Embedding of Datasets} \label{Sec:Embedding}
We construct an embedding-of-datasets: a dense, continuous vector representation of datasets. In this embedding-of-datasets, each dataset is mapped to a single fixed-dimensional vector. Embeddings are typically constructed to compress a high-dimensional vector space into a lower one. Since individual datasets can be of arbitrary dimension, the space of datasets is not a vector space. Therefore, the compression process is nontrivial and cannot directly leverage the numerous well-established techniques developed for standard vector-space compression.

To obtain dataset-level representations, we leverage the internal transformer representations produced by TabPFN during in-context inference. When TabPFN processes a dataset, it embeds each datapoint into a learned feature space as part of its prediction pipeline. We take the mean of these datapoint-level embeddings to aggregate them into a fixed-dimensional vector that summarizes the statistical structure of the entire dataset. We present a learned embedding space of datasets extracted from TabPFN 2.5 in Figure~\ref{fig:embedding}.
%Later, in Section~\ref{Sec:Similarity} we construct a kernel-based embedding using pairwise discrepancies between datasets. 

\subsection{Distinguishability of Datasets} \label{Sec:Distinguishability}
\begin{figure}[!htb]
    \centering
    \includegraphics[width=\linewidth]{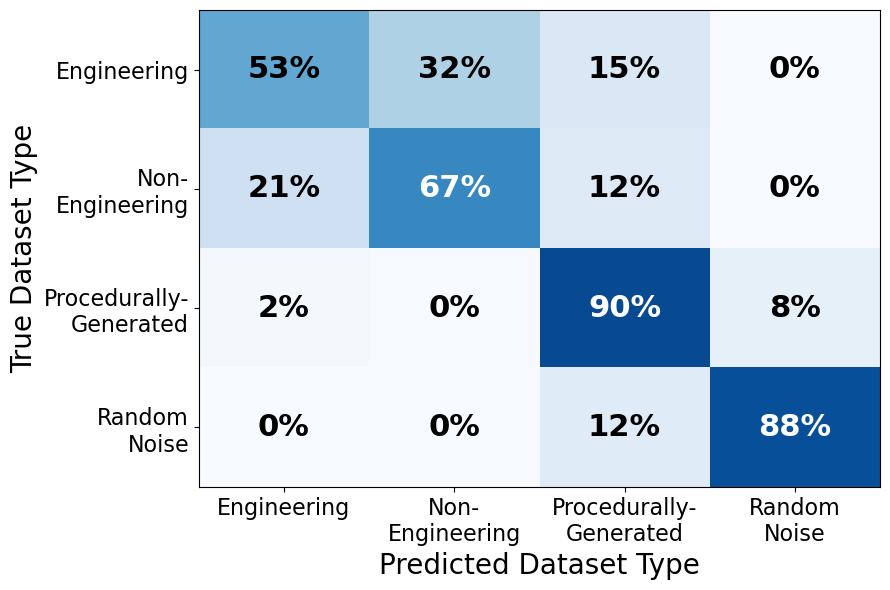}
    \caption{Confusion Matrix illustrating distinguishability of dataset classes (according to TabPFN 2.5's embedding). Distributional differences between engineering and non-engineering datasets can be witnessed. A small portion of the procedurally generated data appears to span the engineering data. }
    \label{fig:confusion}
\end{figure}

We investigate the extent to which different classes of datasets can be classified based on TabPFN 2.5's embedding. We train an XGBoost~\cite{chen2016xgboost} classifier using the hyperparameter optimization process described in the Appendix~\ref{Sec:classifier_optimization}. A confusion matrix of the classification performance is shown in Figure~\ref{fig:confusion}. In binary comparisons, engineering and non-engineering datasets are distinguishable with 65.8\% accuracy. procedurally generated data and engineering data  are distinguishable with 96.8\% accuracy.

\section{A Foundation Model for Engineering Data}
The relatively strong distinguishability between procedurally generated data and engineering data raises a natural question: Can we adapt a purely synthetic training pipeline to better reflect the statistical structure of engineering data without ever training directly on real engineering datasets? To address this question, we leverage the PFN-embedding space introduced in Section~\ref{Sec:Embedding} as a dataset-level representation for distributional comparison.

\subsection{Training Procedure}

\paragraph{Embedding Construction.}
We first compute PFN-based dataset embeddings for 10{,}000 procedurally generated datasets and all 35 engineering datasets. Each dataset is mapped to a 192-dimensional vector by averaging the transformer representations of its datapoints under TabPFN 2.5.

\begin{figure}[!htb]
    \centering
    \includegraphics[width=\linewidth]{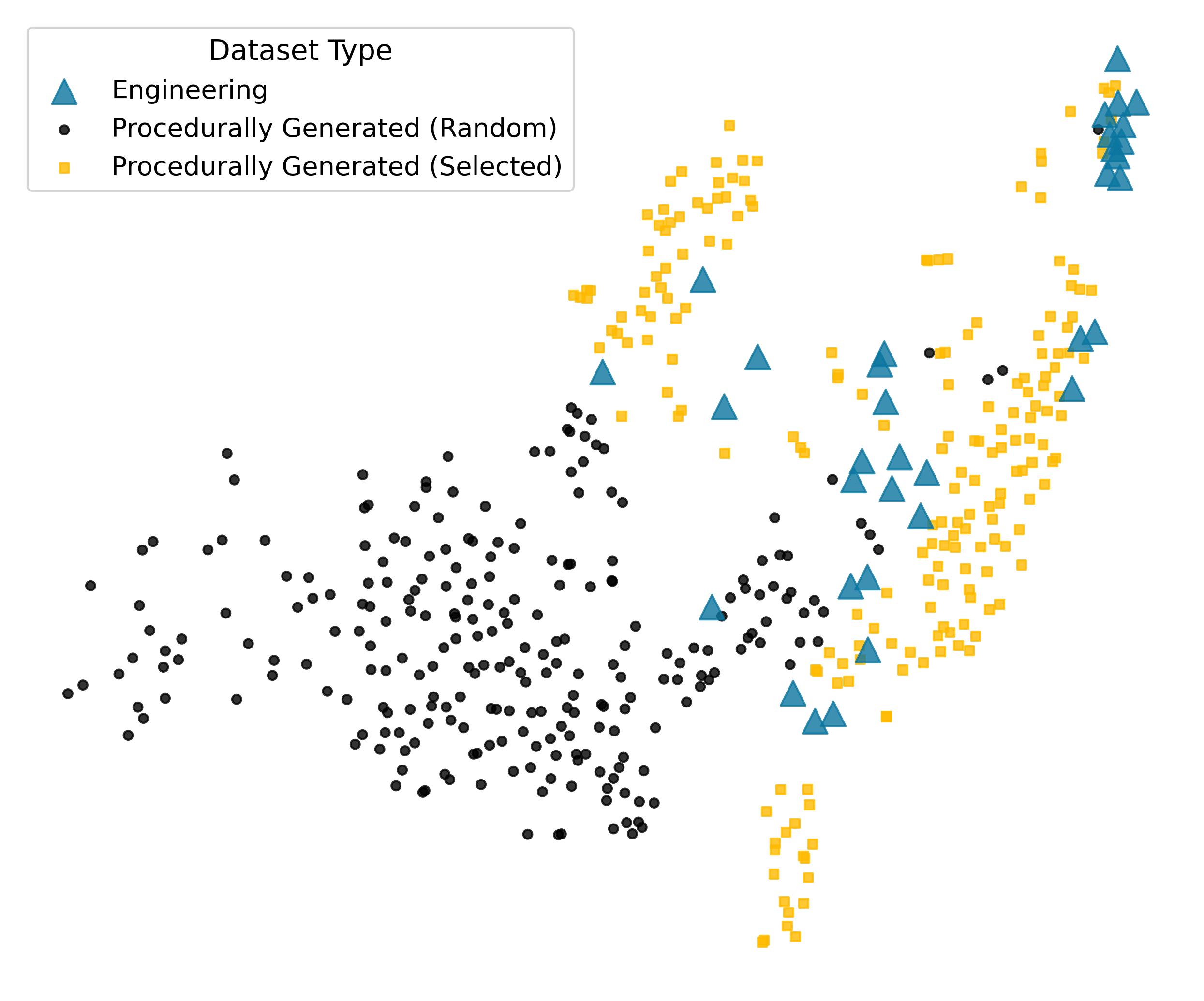}
    \caption{Visualization of procedurally generated data that is subselected to maximally resemble engineering data. Embedding is a 2D t-SNE of the TabPFN 2.5's 192-dimensional dataset embedding. The selected procedurally generated appears to better span the engineering data compared to the randomly generated data.}
    \label{fig:selected_embedding}
\end{figure}

\begin{figure}[!htb]
    \centering
    \includegraphics[width=\linewidth]{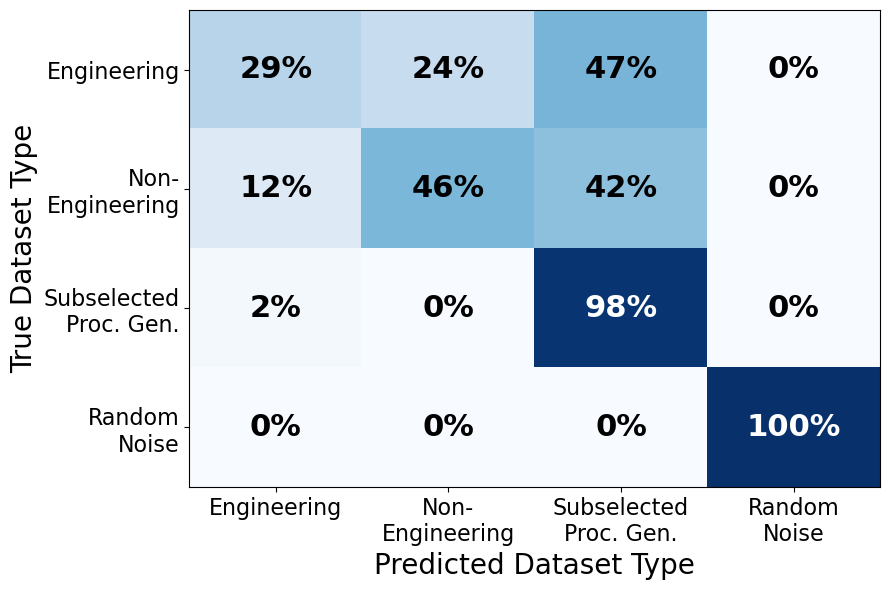}
    \caption{Confusion matrix illustrating that subselected procedurally generated data and engineering data are less easily distinguished (based on TabPFN 2.5's embedding). }
    \label{fig:selected_confusion}
\end{figure}

\paragraph{Selecting Engineering-Like Synthetic Data.}
To identify synthetic datasets that resemble engineering datasets, we frame the problem as binary classification in embedding space. We procedurally generate 10,000 synthetic datasets. We then randomly select 250 procedurally generated dataset embeddings and combine them with embeddings from 70\% of the engineering datasets to train an XGBoost classifier using the hyperparameters determined in Appendix~\ref{Sec:classifier_optimization}. The classifier learns to discriminate between embeddings from engineering datasets and those from procedurally generated datasets.
The classifier is then applied to the remaining 9{,}750 procedurally generated embeddings. For each synthetic dataset, we compute the predicted probability of belonging to the engineering class. We then select the top 200 synthetic datasets with the highest engineering-class probability. As shown in Figure~\ref{fig:selected_embedding}, these subselected datasets appear to span the space of engineering datasets much more thoroughly. This is corroborated through the confusion matrix shown in Figure~\ref{fig:selected_confusion}. The subselected procedurally generated data is only distinguishable from engineering data with 88.9\% accuracy (compared to 96.8\% without subselection).

\paragraph{Fine-Tuning with Selected Synthetic Datasets.}
Following the fine-tuning methodology proposed by~\cite{garg2025real}, we fine-tune TabPFN 2.5 for five epochs using only the 200 selected synthetic datasets. Importantly, no real engineering data are used during fine-tuning; only procedurally generated datasets that most resemble engineering datasets are used.

This procedure can be interpreted as domain adaptation without target-domain supervision: rather than fine-tuning on scarce engineering data, we instead reshape the synthetic training distribution by selecting datasets that are nearest to engineering data in PFN-embedding space. This approach demonstrates that principled sampling of procedurally generated data can bridge domain gaps and improve performance in data-constrained engineering settings.

\subsection{Experimental Results}
\begin{figure*}[!htb]
    \centering
    \includegraphics[width=\linewidth]{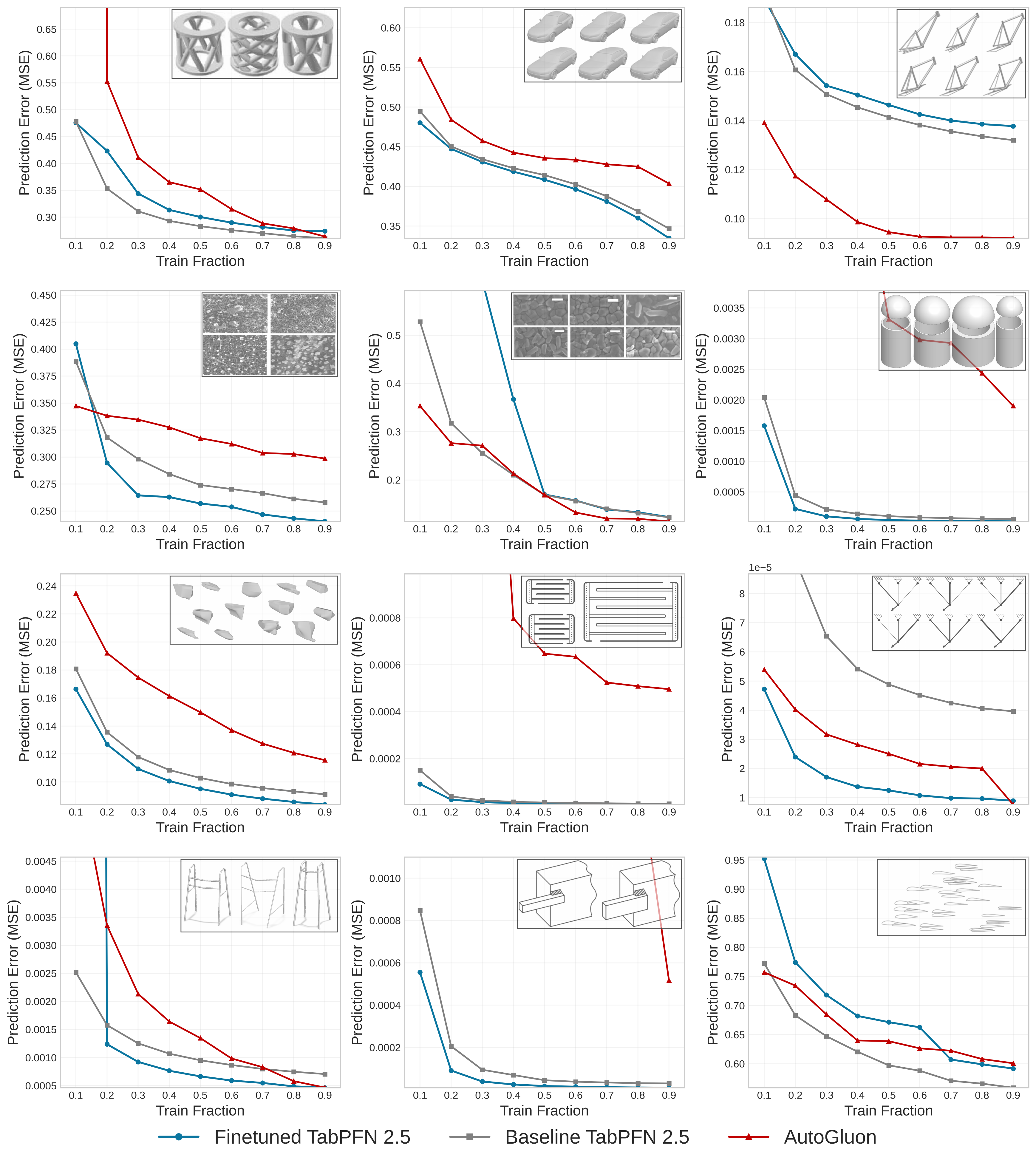}
    \caption{Sample comparison of model accuracy for different quantities of data over 12 example problems. }
    \label{fig:12_comparison}
\end{figure*}

In this section, we discuss our evaluation procedure and key experimental results. 
\paragraph{Downstream Evaluation.}
We evaluate the fine-tuned model on all 35 engineering datasets. For each dataset, we employ a 70\%-30\% train-test split among the datasets used to select synthetic data for model fine-tuning. Model performance is then measured at 9 increments of dataset size (10\%-90\%), calculated during 10-fold cross-validation. This cross-validation scheme ensures that test-time datasets are not utilized in the dataset selection for fine-tuning, nor for inference time context data. Results are compared against (1) the original, non-fine-tuned TabPFN 2.5 model to quantify gains attributable to distribution-aware synthetic data selection, and (2) the state-of-the-art tabular learning model, AutoGluon. As shown in Table~\ref{tab:tabpfn_ft_comparison}, our fine-tuned TabPFN 2.5 model outperforms both its non-fine-tuned counterpart and AutoGluon in 29/35 and 27/35 problems, respectively for a 70\%-30\% train-test split. Comparisons on 12 sample datasets, swept over train size are illustrated in Figure~\ref{fig:12_comparison}. 

\begin{table}[t]
\centering
\caption{Comparison of fine-tuned TabPFN 2.5 against baseline models across 35 engineering datasets using a 70/30 train/test split. Entries report the number and percentage of datasets on which the fine-tuned model achieves lower test error.}
\label{tab:tabpfn_ft_comparison}
\begin{tabular}{lcc}
\toprule
\textbf{Baseline} & \makecell{\textbf{Datasets} \\ \textbf{(/35)}} & \makecell{\textbf{Fraction} \\ \textbf{(\%)}} \\
\midrule
vs.\ TabPFN 2.5 & 29 & 82.9 \\
vs.\ AutoGluon & 27 & 77.1 \\
\bottomrule
\end{tabular}
\end{table}

\paragraph{Data-Efficiency.}

\begin{figure*}[!htb]
    \centering
    \begin{subfigure}{\linewidth}
        \centering
        \includegraphics[height=5cm]{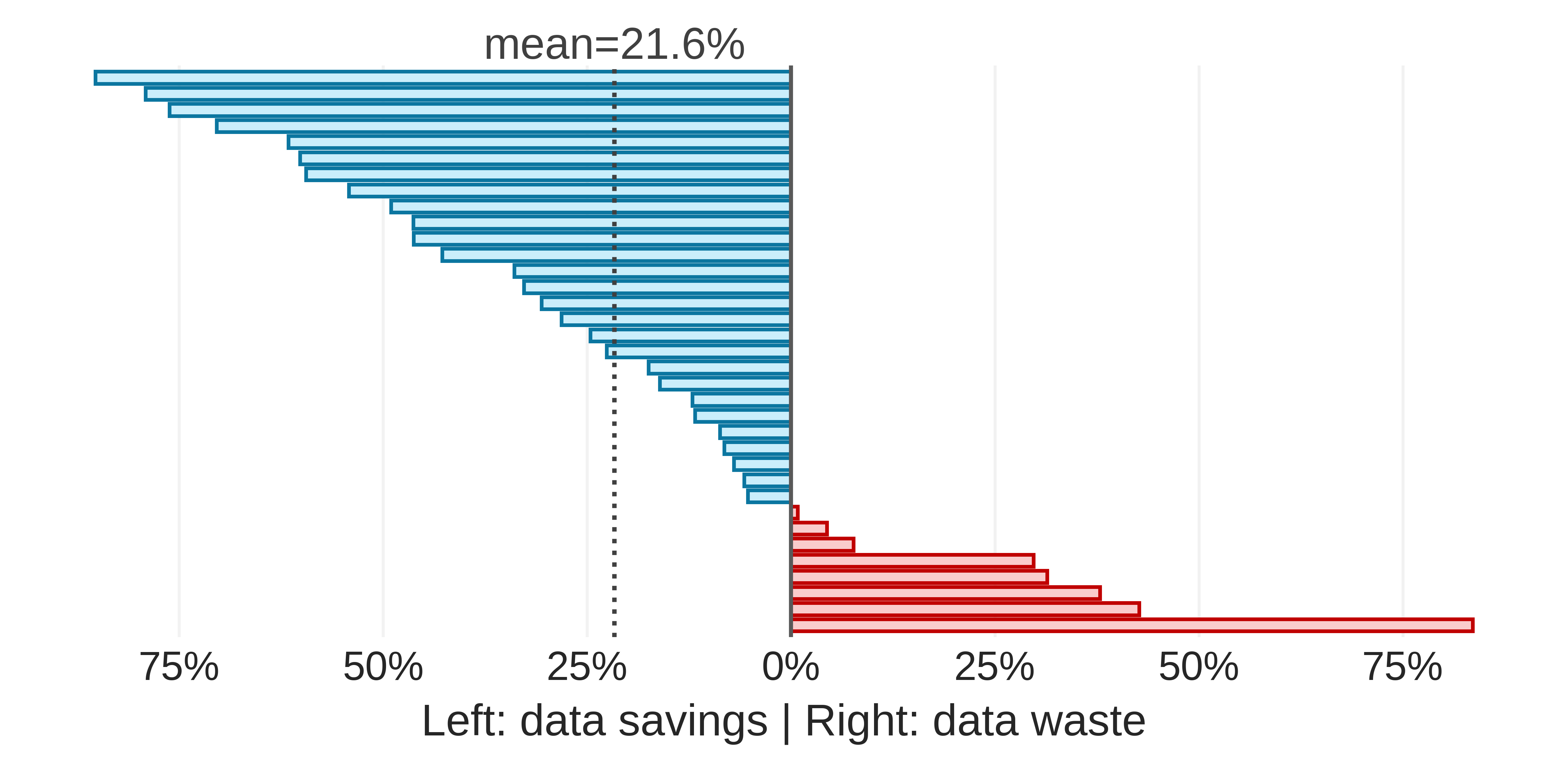}
        \hspace{0.02\linewidth}
        \includegraphics[height=5cm]{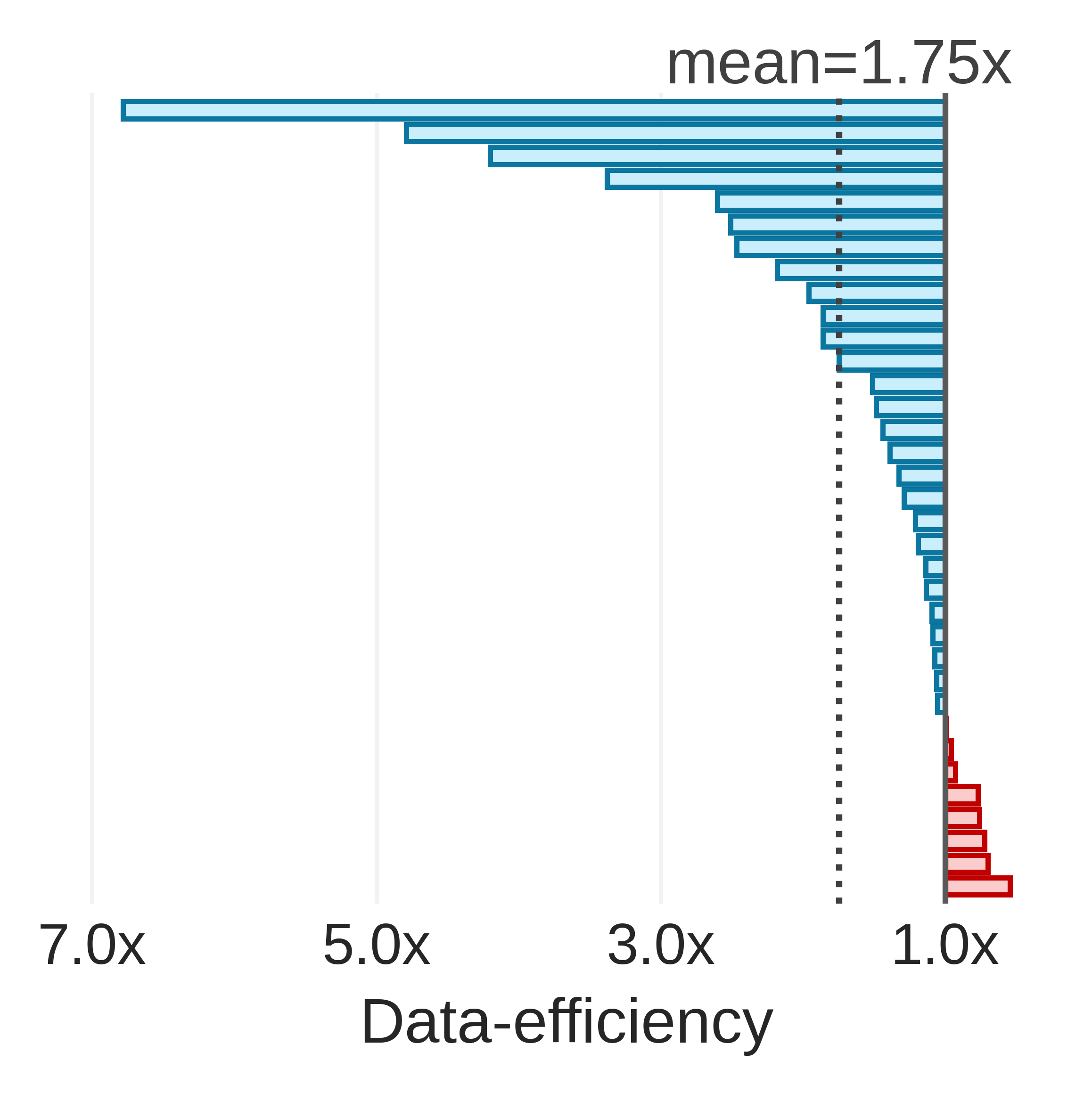}
    \end{subfigure}
    \caption{Comparison of the estimated data-efficiency gain as a result of fine-tuning (our fine-tuned TabPFN 2.5 model vs. non-fine-tuned TabPFN 2.5 model) across all tested datasets. Left: Additive data-efficiency ($E_{add}$). Right: Multiplicative data-efficiency ($E_{mult}$)}
    \label{fig:efficiency_vs_pfn}
\end{figure*}

\begin{figure*}[!htb]
    \centering
    \begin{subfigure}{\linewidth}
        \centering
        \includegraphics[height=5cm]{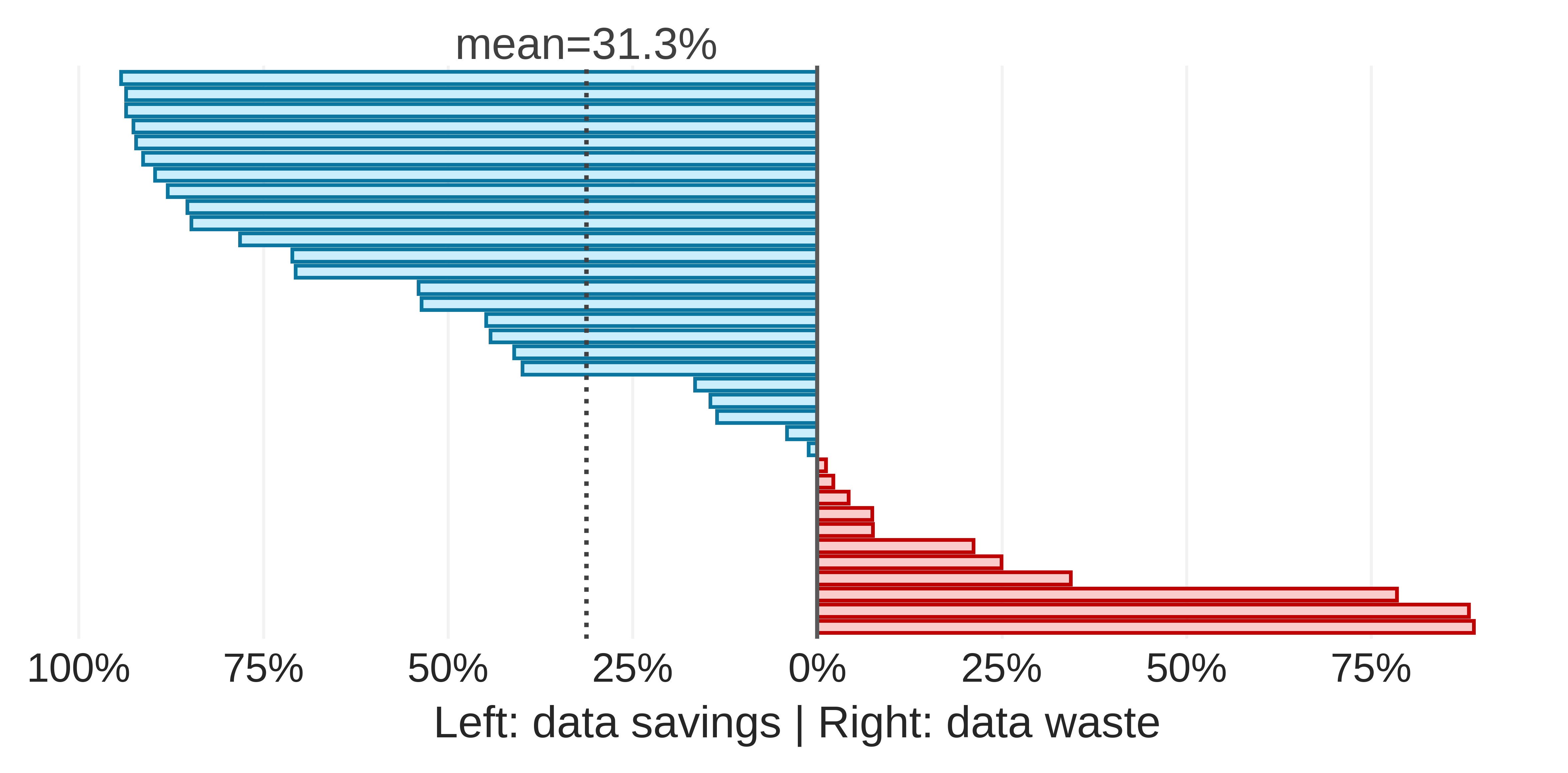}
        \hspace{0.02\linewidth}
        \includegraphics[height=5cm]{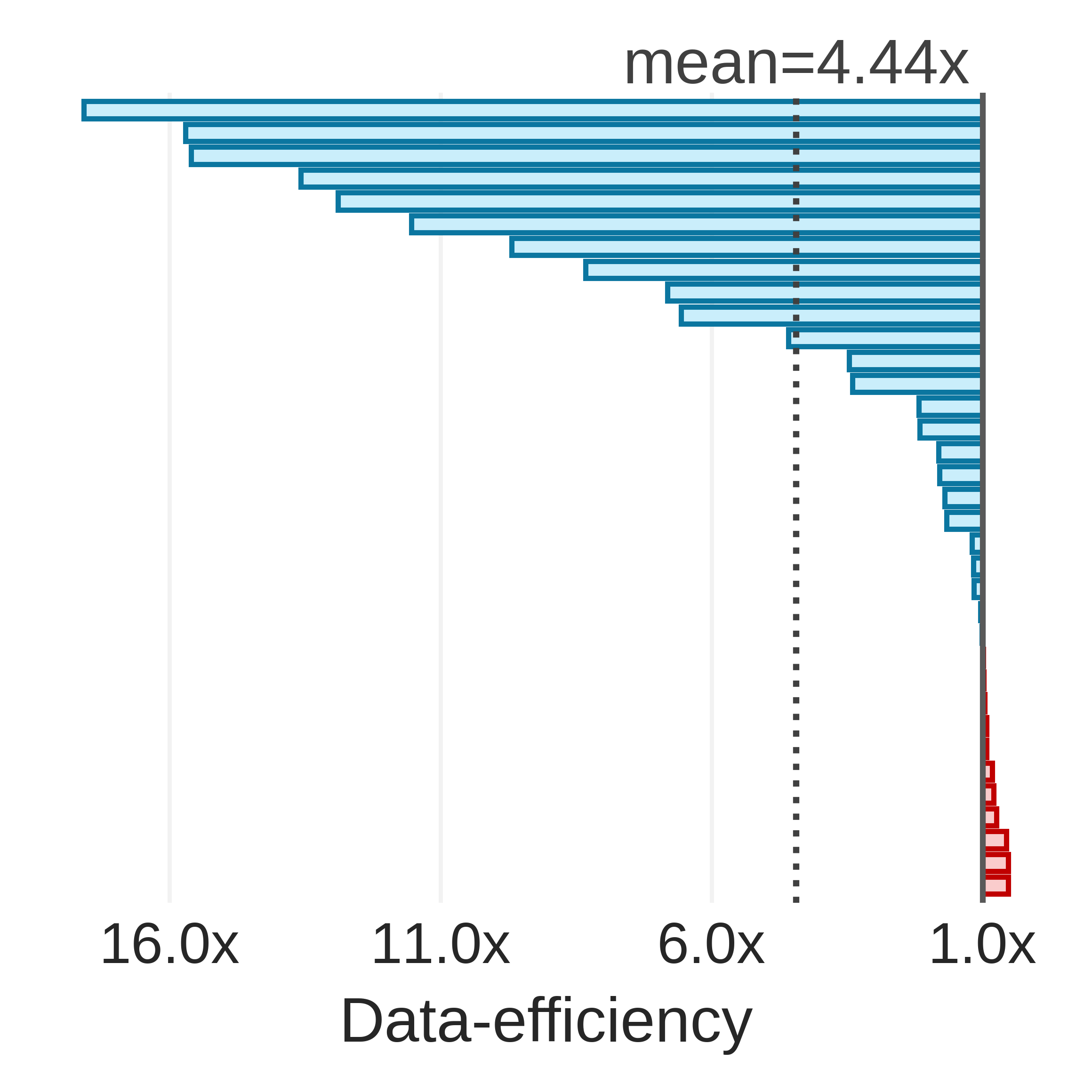}
    \end{subfigure}
    \caption{Comparison of the estimated relative data-efficiency of \textbf{our fine-tuned TabPFN 2.5 model} against \textbf{AutoGluon} across all tested datasets. Left: Additive data-efficiency ($E_{add}$). Right: Multiplicative data-efficiency ($E_{mult}$)}
    \label{fig:efficiency_vs_ag}
\end{figure*}

We measure dataset efficiency by comparing how much training data a \emph{new} model requires to match the performance of a \emph{reference} (ref) model. For each dataset, we first record the reference model’s performance when trained on an amount of data $D_{\text{ref}}$ (taken as $90\%$ of the available training set). We then estimate the amount of data $D_{\text{new}}$ required for the new model to achieve the same mean squared error value by interpolating along its empirical performance curve as a function of training data size. 

We report two complementary measures of data efficiency. The \emph{additive data efficiency} measures the absolute data savings,
\[
E_{\text{add}} = D_{\text{ref}} - D_{\text{new}},
\]
which represents how many fewer training samples the new model requires to reach the same performance. The \emph{multiplicative data efficiency} measures the relative improvement,
\[
E_{\text{mult}} = \frac{D_{\text{ref}}}{D_{\text{new}}},
\]
which indicates how many times more data the reference model needs to match the new model’s efficiency. When $D_{\text{new}} < D_{\text{ref}}$, the new model is more data-efficient; otherwise, the efficiency values become negative (additive) or fall below one (multiplicative), indicating that the reference model requires less data.

If the new model does not reach the reference target performance within the observed data range, we instead anchor the comparison on the new model’s performance at $D_{\text{ref}}$ and estimate how much data the reference model would require to match it by extrapolating along the reference model’s performance curve.

Efficiency gains as a result of the fine-tuning are shown in Figure~\ref{fig:efficiency_vs_pfn}, while an efficiency comparison to AutoGluon is shown in Figure~\ref{fig:efficiency_vs_ag}. 

\subsection{Discussion}
Our experimental results support four key findings. 

\paragraph{Variability in Model Performance.} First, TabPFN 2.5 is a much stronger baseline that AutoGluon on many engineering problems, despite AutoGluon being the leader in broader tabular prediction benchmarks like TabArena~\cite{erickson2025tabarena}. We witness extreme discrepancies in model performance on engineering datasets. This highlights the importance of engineering-focused tabular prediction benchmarks, as general-purpose benchmarks are not representative of engineering performance. 

\paragraph{Fine-tuning on Subselected Synthetic Data Improves Performance.}
In our experiments, our fine-tuning achieves superior performance in a clear (29/35) majority, with average data-efficiency of 1.75x. The advantage is clearly demonstrated for engineering design problems. Given this strong performance, we anticipate that our proposed method of fine-tuning on subselected synthetic data may similarly net performance improvements in other application areas.

\paragraph{Our Fine-Tuned TabPFN 2.5 is State-Of-The-Art for Tabular Engineering Prediction.} 
As noted in~\cite{picard2024untrained}, non-fine-tuned PFNs outcompete baselines in systematic testing on engineering-design problems. Having established the overwhelming improvements over the non-fine-tuned model, we argue that our fine-tuned TabPFN 2.5 model sets the new state-of-the-art for tabular prediction in engineering-related problems. Its strong advantage over AutoGluon, winning 27/35 problems at an average data-efficiency of 4.44x, confirms our model's dominance. 

\paragraph{Performance Gains are Inconsistent.} 
Despite widespread advantages in amortized testing, our model does not consistently outperform the base TabPFN 2.5 nor AutoGluon in every problem, falling short in 6 and 8 problems respectively. Some of these problems see major performance degredation. This highlights the further opportunities to further improve performance and consistency across problems by enhancing not only the subselection of synthetic data, but potentially also the generation procedure itself.

\section{Conclusion}

This work demonstrates that foundation-model paradigms can extend to engineering predictive modeling despite the traditionally small and fragmented nature of engineering datasets. By introducing TREDBench and a dataset-level embedding framework, we provide tools to systematically analyze the statistical structure of engineering data and its relationship to both non-engineering and procedurally generated datasets. Our results show that engineering datasets occupy a partially distinct region of dataset space, yet remain sufficiently similar to a subset of procedurally generated tasks to enable effective transfer through synthetic pretraining.

Leveraging this observation, we propose a principled method for selecting engineering-like synthetic datasets and use them to fine-tune a tabular foundation model without access to real engineering data. The resulting model achieves substantial improvements in predictive performance and data efficiency, outperforming both the base TabPFN 2.5 model and the AutoGluon AutoML system across a majority of the 35 engineering datasets. These results support the broader hypothesis that carefully curated synthetic data can mitigate the data scarcity that has historically limited machine learning adoption in engineering domains.

More broadly, our findings suggest that engineering foundation models may be attainable through improved modeling of the distribution of engineering data. Future work may further refine synthetic data generation pipelines, expand dataset representations beyond tabular regression, and explore larger-scale foundation models specialized for engineering tasks. Such developments could help shift predictive modeling in engineering away from isolated task-specific models toward reusable systems capable of generalizing across diverse design problems.

\bibliographystyle{plainnat} 
\bibliography{biblio}

\appendix
\section{Dataset Details} \label{Sec:dataset_details}
We include a breakdown of datasets included in TREDBench in Table~\ref{tab:tredbench-datasets}. 

\begin{table*}[htb]
\centering
\footnotesize
\begin{minipage}[t]{0.49\textwidth}
\centering
\begin{tabular}{p{0.40\linewidth} l c r}
\toprule
Dataset Name & Src. & Type & Dimensions \\
\midrule
a1 & \cite{nejjar2024context} & NE & 198 x 11 \\
a2 & \cite{nejjar2024context} & NE & 198 x 11 \\
a3 & \cite{nejjar2024context} & NE & 198 x 11 \\
a4 & \cite{nejjar2024context} & NE & 198 x 11 \\
a5 & \cite{nejjar2024context} & NE & 198 x 11 \\
a6 & \cite{nejjar2024context} & NE & 198 x 11 \\
a7 & \cite{nejjar2024context} & NE & 198 x 11 \\
AgNP & \cite{liang2021benchmarking} & E & 3,295 x 5 \\
Ailerons & \cite{grinsztajn2022tree} & E & 13,750 x 33 \\
airfoils & \cite{picard2024untrained} & E & 1,108 x 100 \\
analcatdata\_supreme & \cite{grinsztajn2022tree} & NE & 4,052 x 7 \\
auction\_verification & \cite{fischer2023openml} & NE & 2,043 x 7 \\
AutoAM\_Score & \cite{liang2021benchmarking} & E & 100 x 4 \\
Bike\_Sharing\_Demand & \cite{grinsztajn2022tree} & NE & 17,379 x 6 \\
black\_friday & \cite{gijsbers2024amlb} & NE & 166,821 x 9 \\
brazilian\_houses & \cite{fischer2023openml} & NE & 10,692 x 9 \\
CantileverBeam & \cite{yu2025fast} & E & 1,024 x 21 \\
Car & \cite{yu2025fast} & E & 1,024 x 21 \\
cars & \cite{fischer2023openml} & NE & 804 x 17 \\
class & \cite{nejjar2024context} & NE & 167 x 4 \\
CompressionSpring & \cite{yu2025fast} & E & 1,024 x 7 \\
concrete\_compress... & \cite{fischer2023openml} & E & 1,030 x 8 \\
cps88wages & \cite{fischer2023openml} & NE & 28,155 x 6 \\
cpu\_activity & \cite{fischer2023openml} & NE & 8,192 x 21 \\
Crossed\_barrel\_to... & \cite{liang2021benchmarking} & E & 1,800 x 4 \\
diamonds & \cite{fischer2023openml} & NE & 53,940 x 9 \\
DrivAerNet & \cite{elrefaie2024drivaernet} & E & 3,747 x 23 \\
elevators & \cite{grinsztajn2022tree} & E & 16,599 x 16 \\
energy\_efficiency & \cite{fischer2023openml} & E & 768 x 8 \\
fifa & \cite{fischer2023openml} & NE & 19,178 x 28 \\
forest\_fires & \cite{fischer2023openml} & NE & 517 x 12 \\
FRAMED & \cite{picard2024untrained} & E & 4,800 x 37 \\
grid\_stability & \cite{fischer2023openml} & E & 10,000 x 12 \\
health\_insurance & \cite{fischer2023openml} & NE & 22,272 x 11 \\
HeatExchanger & \cite{yu2025fast} & E & 1,024 x 14 \\
house\_16H & \cite{grinsztajn2022tree} & NE & 22,784 x 16 \\
house\_sales & \cite{grinsztajn2022tree} & NE & 21,613 x 15 \\
houses & \cite{grinsztajn2022tree} & NE & 20,640 x 8 \\
HousValue & \cite{nejjar2024context} & NE & 506 x 13 \\
Human Preference & \cite{regenwetter2025bikebench} & E & 200 x 3 \\
kin8nm & \cite{fischer2023openml} & E & 8,192 x 8 \\
medical\_charges & \cite{grinsztajn2022tree} & NE & 163,065 x 3 \\
miami\_housing & \cite{fischer2023openml} & NE & 13,932 x 15 \\
\bottomrule
\end{tabular}
\end{minipage}
\hfill
\begin{minipage}[t]{0.49\textwidth}
\centering
\begin{tabular}{p{0.40\linewidth} l c r}
\toprule
Dataset Name & Src. & Type & Dimensions \\
\midrule
Moneyball & \cite{fischer2023openml} & NE & 1,232 x 14 \\
naval\_propulsion\_... & \cite{fischer2023openml} & E & 11,934 x 14 \\
nyc-taxi-green-de... & \cite{grinsztajn2022tree} & NE & 581,835 x 9 \\
OnlineNewsPopularity & \cite{gijsbers2024amlb} & NE & 39,644 x 59 \\
P3HT\_conductivity & \cite{liang2021benchmarking} & E & 233 x 5 \\
particulate-matte... & \cite{grinsztajn2022tree} & NE & 394,299 x 6 \\
Perovskite\_Instab... & \cite{liang2021benchmarking} & E & 139 x 3 \\
physiochemical\_pr... & \cite{fischer2023openml} & E & 45,730 x 9 \\
pol & \cite{grinsztajn2022tree} & NE & 15,000 x 26 \\
PressureVessel & \cite{yu2025fast} & E & 1,024 x 8 \\
pumadyn32nh & \cite{fischer2023openml} & E & 8,192 x 32 \\
QSAR\_fish\_toxicity & \cite{fischer2023openml} & NE & 908 x 6 \\
quake & \cite{gijsbers2024amlb} & NE & 2,178 x 3 \\
red\_wine & \cite{fischer2023openml} & NE & 1,599 x 11 \\
ReinforcedConcret... & \cite{yu2025fast} & E & 1,024 x 5 \\
rej & \cite{nejjar2024context} & NE & 4,499 x 8 \\
Rings & \cite{nejjar2024context} & NE & 4,177 x 8 \\
sarcos & \cite{fischer2023openml} & E & 48,933 x 21 \\
ScaledSoundPressure & \cite{nejjar2024context} & NE & 1,503 x 5 \\
seattlecrime6 & \cite{grinsztajn2022tree} & NE & 52,031 x 4 \\
sensory & \cite{gijsbers2024amlb} & NE & 576 x 11 \\
SGEMM\_GPU\_kernel\_... & \cite{grinsztajn2022tree} & NE & 241,600 x 9 \\
ShipD C Set 1 & \cite{narayanan2023data} & E & 10,000 x 44 \\
socmob & \cite{fischer2023openml} & NE & 1,156 x 5 \\
Solar Heat Exchanger & \cite{picard2024untrained} & E & 500 x 2 \\
solar\_flare & \cite{fischer2023openml} & NE & 1,066 x 10 \\
space\_ga & \cite{fischer2023openml} & NE & 3,107 x 6 \\
SpeedReducer & \cite{yu2025fast} & E & 1,024 x 16 \\
student\_performan... & \cite{fischer2023openml} & NE & 649 x 30 \\
sulfur & \cite{grinsztajn2022tree} & E & 10,081 x 6 \\
superconductivity & \cite{fischer2023openml} & E & 21,263 x 81 \\
topo\_2\_1 & \cite{grinsztajn2022tree} & E & 8,885 x 255 \\
Truss 6D & \cite{picard2024untrained} & E & 100,000 x 6 \\
video\_transcoding & \cite{fischer2023openml} & NE & 68,784 x 18 \\
visualizing\_soil & \cite{grinsztajn2022tree} & NE & 8,641 x 4 \\
Walker & \cite{narayanan2023data} & E & 5,007 x 16 \\
wave\_energy & \cite{fischer2023openml} & NE & 72,000 x 48 \\
Welded Beam 2 & \cite{picard2024untrained} & E & 2,000 x 4 \\
WeldedBeam & \cite{yu2025fast} & E & 1,024 x 9 \\
white\_wine & \cite{fischer2023openml} & NE & 4,898 x 11 \\
wine\_quality & \cite{grinsztajn2022tree} & NE & 6,497 x 11 \\
yprop\_4\_1 & \cite{grinsztajn2022tree} & E & 8,885 x 42 \\
 &  &  &  \\
\bottomrule
\end{tabular}
\end{minipage}
\caption{TREDBench dataset summary showing name, source, classification, and dimensionality.}
\label{tab:tredbench-datasets}
\end{table*}
\section{Implementation Details}
\subsection{Hyperparameter Optimization for Dataset Classifiers Trained on TabPFN 2.5 Embedding} \label{Sec:classifier_optimization}

We tuned a multiclass XGBoost classifier using Optuna~\cite{optuna_2019} with a five-fold stratified cross-validation objective maximizing mean balanced accuracy. The search space is reported in Table~\ref{tab:xgb_optuna_search}. Optimization was performed for 50 trials, initialized from a reasonable hand-tuned configuration. The best-performing configuration selected by Optuna is also shown in Table~\ref{tab:xgb_optuna_search}. All other training settings followed standard XGBoost defaults, with the histogram-based tree method and a multiclass soft-probability objective.

\begin{table}[!htb]
\centering
\caption{Optuna search space and selected hyperparameters for multiclass XGBoost trained on embedding features. $n\_estimators$ was considered in intervals of 100}
\label{tab:xgb_optuna_search}
\resizebox{\columnwidth}{!}{
\begin{tabular}{llll}
\toprule
\textbf{Parameter} & \textbf{Domain} & \textbf{Scale} & \textbf{Best} \\
\midrule
$n\_estimators$      & $[300,2000]$ & Linear & 1700 \\
$learning\_rate$     & $[0.01,0.2]$            & Log    & 0.0158 \\
$max\_depth$         & $[3,10]$                & Linear & 7 \\
$subsample$          & $[0.6,1.0]$             & Linear & 0.903 \\
$colsample\_bytree$  & $[0.6,1.0]$             & Linear & 0.622 \\
$min\_child\_weight$ & $[1,20]$                & Log    & 3.84 \\
$reg\_lambda$        & $[10^{-3},50]$          & Log    & 3.12 \\
$gamma$              & $[0,2]$                 & Linear & 0.630 \\
$reg\_alpha$         & $[10^{-4},10]$          & Log    & $0.00406$ \\
\bottomrule
\end{tabular}
}
\end{table}

% \subsection{Ablating PCA Dimension for Classifiers Trained on Inter-Dataset Distance} \label{Sec:PCA_ablate}
% We ablate the selection of the PCA dimension in Figure~\ref{fig:pca_ablate}, finding the optimal dimension to be 20. 
% \begin{figure}[!htb]
%     \centering
%     \includegraphics[width=\linewidth]{images/pca_ablate.png}
%     \caption{Ablation: PCA Dimension for optimal classifier using flattened inter-dataset distance features.}
%     \label{fig:pca_ablate}
% \end{figure}

\end{document}